\documentclass[11pt,a4paper]{article}
\usepackage{authblk}
\usepackage[hyperref]{naaclhlt2019}
\usepackage{times}
\usepackage{latexsym}
\usepackage{amssymb}
\usepackage{amsmath}
\usepackage{amssymb}
\usepackage{graphicx}
\usepackage{url}

\aclfinalcopy 

\title{Ghmerti at SemEval-2019 Task 6: A Deep Word- and Character-based Approach to Offensive Language Identification}

\author[$\clubsuit,\spadesuit$]{Ehsan Doostmohammadi}
\author[$\clubsuit$]{Hossein Sameti}
\author[$\spadesuit$]{Ali Saffar}
\affil[$\clubsuit$]{Speech Processing Lab, Department of Computer Engineering,}
\affil[ ]{Sharif University of Technology, Tehran, Iran}
\affil[$\spadesuit$]{NazarBin, Tehran, Iran}
\affil[ ]{\tt{e.doostm72@student.sharif.edu, sameti@sharif.edu,}}
\affil[ ]{\tt{saffar@nazarbin.com}}

\date{}

\begin{document}
\maketitle
\begin{abstract}
This paper presents the models submitted by Ghmerti team for subtasks A and B of the OffensEval shared task at SemEval 2019. OffensEval addresses the problem of identifying and categorizing offensive language in social media in three subtasks; whether or not a content is offensive (subtask A), whether it is targeted (subtask B) towards an individual, a group, or other entities (subtask C). The proposed approach includes character-level Convolutional Neural Network, word-level Recurrent Neural Network, and some preprocessing. The performance achieved by the proposed model for subtask A is 77.93\% macro-averaged F\textsubscript{1}-score.
\end{abstract}

\section{Introduction}
\label{sec:intro}

The massive rise in user-generated web content, alongside with the freedom of speech in social media and anonymity of the users has brought about an increase in online offensive content and anti-social behavior. The consequences of such behavior on genuine users of the social media have become a serious concern for researchers in Natural Language Processing and related fields in recent years.

The shared task number 6 at SemEval 2019, OffensEval \cite{offenseval}, proposes to model the task of offensive language identification hierarchically, which means identifying the offensive content, whether it is targeted, and if so, the target of the offense.
In OffensEval, offensive language is defined as ``any form of non-acceptable language (profanity) or a targeted offense, which can be
veiled or direct'' which includes ``insults,
threats, and posts containing profane language or
swear words'' \cite{offenseval}.

We have participated in the first two subtasks (A and B) of OffensEval with the proposed approach of a deep model consisting of a Recurrent Neural Network (RNN) for word-level and Convolutional Neural Network (CNN) for character-level processing\footnote{You can find the code of the deep model on this project's repository on github: \href{https://github.com/edoost/offenseval}{github.com/edoost/offenseval}}. Character-level processing is beneficial, as offensive comments are likely to follow unorthodox writing styles, contain obfuscated words, or have irregular word separation which leads to tokenization issues \cite{mehdad2016characters,nobata2016abusive}. We also experimented with two other methods, a Support Vector Machine (SVM) with TFIDF and count features and another SVM with BERT \cite{devlin2018bert} -encoded sentences as input, both with lower performances comparing with the deep model.

After overviewing the related work in section \ref{sec:relwork}, we discuss the methodology and the data in details in section \ref{sec:meth}, and the results in section \ref{sec:results}. In section \ref{sec:analysis}, we analyze the results and conclude the paper in section \ref{sec:conc}.

\section{Related Work}
\label{sec:relwork}

Offensive language identification has been of interest for researchers in recent years. Early work in the related fields include detection of online trolling \cite{cambria2010not}, racism \cite{greevy2004classifying}, and cyberbullying \cite{dinakar2012common}.

Papers published in recent years include  \cite{davidson2017automated}, which introduces the Hate Speech Detection dataset and experiments with different machine learning models, such as logistic regression, na\"{i}ve Bayes, random forests, and linear SVMs to investigate hate speech and offensive language, \cite{malmasi2017detecting} which experiments further on the same dataset using SVMs with n-grams and skip-grams features, and \cite{gamback2017using} and \cite{zhang2018detecting}, both exploring the performance of neural networks and comparing them with other machine learning approaches.
Also, there has been published a couple of surveys covering various work addressing the identification of abusive, toxic, and offensive language, hate speech, etc., and their methodology including \cite{schmidt2017survey} and \cite{fortuna2018survey}.

Additionally, there were several workshops and shared tasks on offensive language identification and related problems, including TA-COS\footnote{\url{http://ta-cos.org/}}, Abusive Language Online\footnote{\url{https://sites.google.com/site/abusivelanguageworkshop2017/}}, and TRAC\footnote{\url{https://sites.google.com/view/trac1/home}}\cite{trac2018report}, and GermEval \cite{wiegand2018overview}, which shows the significance of the problem.

\section{Methodology and Data}
\label{sec:meth}

The methodology used for both subtask A, offensive language identification, and subtask B, automatic categorization of offense types, consists of a preprocessing phase and a deep classification phase. We first introduce the preprocessing phase, then elaborate on the classification phase.

\subsection{Preprocessing}
The preprocessing phase consists of (1) replacing obfuscated offensive words with their correct form and (2) tweet tokenization using NLTK tweet tokenizer \cite{bird2009natural}. In social media, some words are distorted in a way to escape the offense detection systems or to reduce the impertinence. For instance, `asshole' may be written as `a\$\$hole', `a\$sh0le', `a**hole', etc. Having a list of English offensive words, we can create a list containing most of the possible permutations. 
Using such a list will ease the job for the classifier and searching in it is computationally cheap. Furthermore, replacing contractions, e.g. `I'm' with `I am', and replacing common social media abbreviations, e.g. `w/' with `with', were not helpful and were not used to train the final model.

\subsection{Deep Classifier}
\label{subsec:deep}

Given a tweet, we want to know if its offensive or not (subtask A), and if the offense is targeted (subtask B). Regarding that both subtasks are problems of binary classification, we used one architecture to tackle both. To define the problem, if we have a tweet $x$, we want to predict the label $y$, \tt OFF \normalfont or \tt NOT \normalfont in subtask A, and \tt TIN \normalfont or \tt UNT \normalfont in subtask B. 
Two representations are therefore created for each input $x$:

\begin{enumerate}
    \item $x_c$ which is the indexed representation of the tweet based on its characters padded to the length of the longest word in the corpus. The indices include 256 of the most common characters, plus 0 for padding and 1 for unknown characters. 
    \item $x_w$ which is the embeddings of the words in the input tweet based on FastText's 600B-token common crawl model \cite{mikolov2018advances}.
\end{enumerate}

Then, $x_c$ is fed into an embedding layer with output size of 32 and a CNN layer after that. $x_c$ is then concatenated with $x_w$ and both are fed to a unidirectional RNN with LSTM cell of size 256, the output of which is the input to two consecutive fully-connected layers that map their input to an $\mathbb{R}^{128}$ and an $\mathbb{R}^2$ space, respectively.
We also applied dropout of keeping rate $0.5$ on CNN's output, $x_w$, RNN's output, and the first fully-connected layer's output.

The CNN layer consists of four consecutive sub-layers:

\begin{enumerate}
    \item CNN consisting of 64 filters with kernel size of 2, stride of 1, same padding and RELU activation;
    \item max-pooling layer with pool size and stride of 2;
    \item another CNN, same as the first one, but with 128 filters;
    \item the same max-pooling again.
\end{enumerate}

Finally, we used an AdamOptimizer \cite{kingma2014adam} with learning rate of $1\mathrm{e}{-3}$ and batch size of 32 to train the model.

\subsection{Baseline Methods}
\label{subsec:baseline}

We used two baseline methods for subtask A:

\begin{itemize}
    \item an SVM with 1- to 3-gram word TFIDF and 1- to 5-gram character count feutrue vectors as input;
    
    \item an SVM with BERT representations of the tweets (using average pooling \cite{xiao2018bertservice}) as input using \tt BERT-Large, Uncased \normalfont model. 
\end{itemize}

The SVMs were trained for 15 epochs with stochastic gradient descent, hinge loss, alpha of $1\mathrm{e}{-6}$, \tt elasticnet \normalfont penalty, and \tt random\_state \normalfont of 5. The SVMs were implemented using Scikit-learn \cite{scikit-learn}.




\subsection{Data}
\label{subsec:data}

The main dataset used to train the model is Offensive Language Identification Dataset (OLID) \newcite{OLID}. The dataset is annotated hierarchically to identify offensive language (OFFensive or NOT), whether it is targeted (Targeted INsult or UNTargeted), and if so, its target (INDividual, GRouP, or OTHer).
We divided the 13,240 samples in the training set into 12,000 samples for training and 1,240 samples for validation. 

As neural networks require huge amount of training data, we tried adding more data from the dataset of the First Workshop on Trolling, Aggression, and Cyberbullying (TRAC-1) \cite{trac2018report} which was not helpful. 
However, adding the training data from Toxic Comment Classification Challenge on Kaggle \cite{jigsawkaggle} increased the macro-averaged F\textsubscript{1}-score on the validation set by ${\thicksim2\%}$. This data comprises tweets with positive and negative tags in six categories: 
\tt toxic\normalfont,
\tt severe\_toxic\normalfont,
\tt obscene\normalfont,
\tt threat\normalfont,
\tt insult\normalfont,
\tt identity\_hate\normalfont.
We only used \tt toxic \normalfont and \tt severe\_toxic \normalfont positive samples as \tt{OFF} \normalfont and the ones with no positive label in any category as \tt{NOT}\normalfont. None of the data from other categories, either positive or negative, were included in the additional training data. 
After that, we were left with 109,236 samples, most of which were labeled as \tt NOT\normalfont. To balance 
\tt OFF \normalfont and \tt NOT \normalfont samples, 84,626 of \tt NOT \normalfont samples were randomly removed. 
In the end, 12,305 \tt OFF \normalfont and 12,305 \tt NOT \normalfont samples were added to the training data.

\section{Results}
\label{sec:results}

Finally, we trained the baseline models in \ref{subsec:baseline} and the model described in \ref{subsec:deep} using the combination of the OLID training data and the data from Toxic Comment Classification Challenge (which is described in \ref{subsec:data}).

You can see the macro-averaged F\textsubscript{1}-score and accuracy on the test set for the baseline scores provided by task organizers, baseline methods we used (on both training and validation data), and the deep classifier model (DeepModel) in table \ref{tab:results-A-open}. DeepModel is trained on the training data (not including the validation data) and DeepModel+val on the combination of the training and validation data. The best performance is in bold.

\begin{table}[h]
\center
\begin{tabular}{l|cc}
\hline
\hline
\bf System & \bf Macro F\textsubscript{1} & \bf Accuracy \\ 
\hline
All NOT baseline & 0.4189 & 0.7209 \\
All OFF baseline & 0.2182 & 0.2790 \\
\hline
SVM & 0.7452 & 0.8011 \\
BERT-SVM & 0.7507 & 0.8011 \\
\hline
DeepModel & 0.7788 & 0.8326 \\
\bf{DeepModel+val} & \bf{0.7793} & \bf{0.8337} \\
\hline
\hline
\end{tabular}
\caption{Results for subtask A}
\label{tab:results-A-open}
\end{table}

\begin{figure}[h]
\centering
\includegraphics[width=0.5\textwidth]{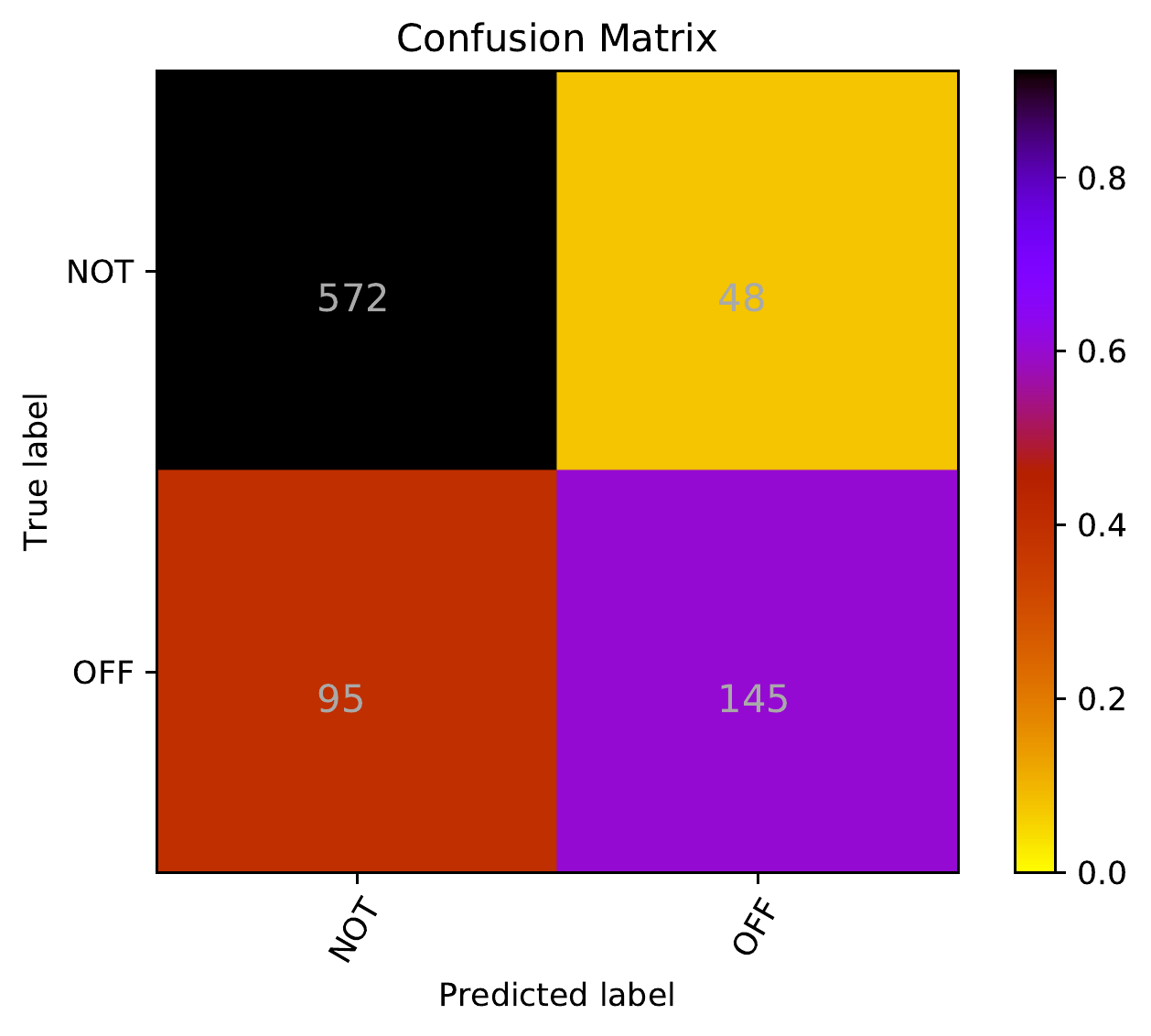}
\caption{The confusion matrix for DeepModel+val in subtask A}
\label{fig:1}
\end{figure}

The best performance belongs to DeepModel+val by a margin of more than 2.8 percent, with the best baseline performance, BERT-SVM. However, it should be mentioned that the results in the first two rows belong to a model trained only on OLID. You can see the confusion matrix for the best performance in figure \ref{fig:1}.

From the confusion matrix we can see that the performance of DeepModel+val on \tt NOT \normalfont is quite good, but not on \tt OFF\normalfont. You can see the detailed results of DeepModel+val in table \ref{tab:results-A-details}.

\begin{table}[h]
\center
\begin{tabular}{c|ccc}
\hline
\hline
& \bf Precision & \bf Recall & \bf F\textsubscript{1}-score \\
\hline
\tt NOT & 0.8576 & 0.9226 & 0.8889 \\
\tt OFF & 0.7513 & 0.6042 & 0.6697 \\
\hline
\hline
\end{tabular}
\caption{Detailed DeepModel+val results in subtask A}
\label{tab:results-A-details}
\end{table}

In subtask B, DeepModel+val outperformed the baseline results by a large margin, like subtask A. The results for subtask B are presented in table \ref{tab:results-B-open}.

\begin{table}[h]
\center
\begin{tabular}{l|cc}
\hline
\hline
\bf System & \bf Macro F\textsubscript{1} & \bf Accuracy \\ 
\hline
All TIN baseline & 0.4702 & 0.8875 \\
All UNT baseline & 0.1011 & 0.1125 \\
\hline
DeepModel & 0.6065 & 0.8583 \\
\bf{DeepModel+val} & \bf{0.6400} & \bf{0.8875} \\
\hline
\hline
\end{tabular}
\caption{Results for subtask B}
\label{tab:results-B-open}
\end{table}

\begin{figure}[h]
\centering
\includegraphics[width=0.5\textwidth]{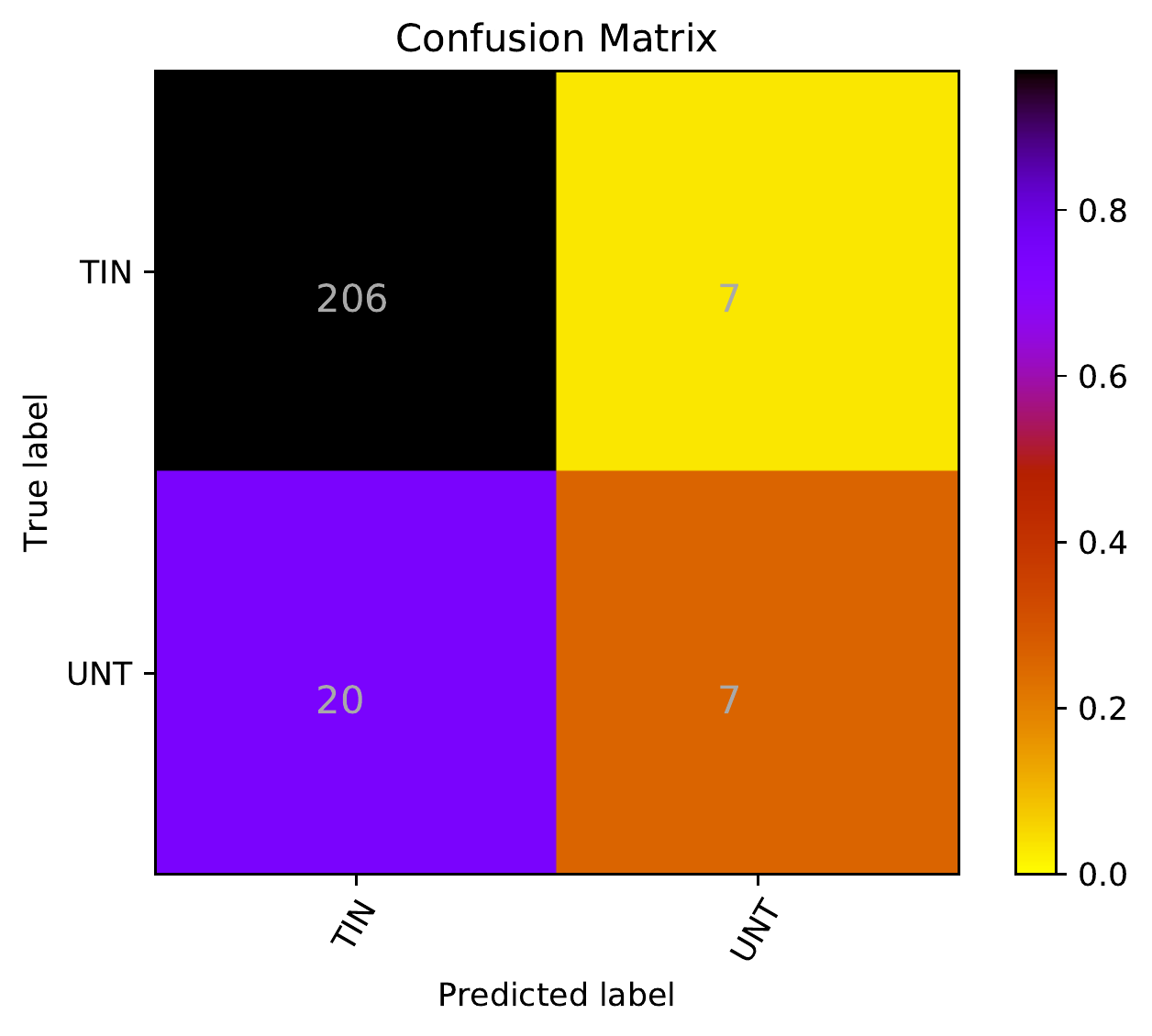}
\caption{The confusion matrix for the DeepModel+val in subtask B}
\label{fig:2}
\end{figure}

This time, adding the validation data made a considerable difference, as the training data for subtask B is fewer. You can see the confusion matrix for DeepModel+val in figure \ref{fig:2}.

The confusion matrix shows that the performance of the model is good for \tt TIN\normalfont, but poor for \tt UNT\normalfont. Table \ref{tab:results-B-details} shows the detailed results for DeepModel+val in subtask B, which indicates that the imbalance is worse than subtask A and the poor performance on \tt UNT \normalfont is mainly due to low recall.

\begin{table}[h]
\center
\begin{tabular}{c|ccc}
\hline
\hline
& \bf Precision & \bf Recall & \bf F\textsubscript{1}-score \\
\hline
\tt TIN & 0.9115 & 0.9671 & 0.9385 \\
\tt UNT & 0.5000 & 0.2593 & 0.3415 \\
\hline
\hline
\end{tabular}
\caption{Detailed DeepModel+val results in subtask B}
\label{tab:results-B-details}
\end{table}

\section{Analysis}
\label{sec:analysis}
In subtask A, DeepModel+val outperformed the second best method, BERT-SVM, by 2.86\% Macro F\textsubscript{1}-score. BERT-SVM results, however, were not much better than the SVM with TFIDF and count features, probably due the fact that the BERT model requires fine-tuning for more task-specific representations.

The majority of DeepModel+val's errors are in \tt OFF \normalfont class and can be categorized into 
(1) sarcasm: the model is unable to detect sarcastic language which is even difficult for humans to detect; 
(2) emotion: discerning emotions, such as anger, seems to be a challenge for the model;
(3) ethnic and racial slurs, etc. Solving these problems require a more comprehensive knowledge of the context and the language, which was examined in works such as \cite{poria2016deeper} and improved the results. However, experimenting with emotion embeddings in the current work was not helpful and did not appear in the final results. Being aware of the emotion of the text, personality of the author, and sentiment of the sentences is helpful to detect offensive language, as many offensive contents have an angry tone \cite{elsherief2018hate} or do not contain profane language \cite{malmasi2018challenges}. One can also make use of the benefits of BERT's context and sentence sequence awareness by fine-tuning it on the training data, which is computationally expensive and was not feasible for the authors of this paper.

\section{Conclusion}
\label{sec:conc}

In this paper, we introduced Ghmerti team's approach to the problems of `offensive language identification' and `automatic categorization of offense type' in shared task 6 of SemEval 2019, OffensEval. In subtask A, the neural network-based model outperformed the other methods, including an SVM with word TFIDF and character count features and another SVM with BERT-encoded tweets as input. Furthermore, analysis of the results indicates that sarcastic language, inability to discern the emotions such as anger, and ethnic and racial slurs constitute a considerable portion of the errors. Such deficiencies demand larger training corpora and variety of other features, such as information on sarcasm, emotion, personality, etc. 

\bibliography{semeval}
\bibliographystyle{acl_natbib}

\end{document}